\documentclass{article}

\PassOptionsToPackage{numbers, compress}{natbib}

\usepackage[final]{neurips_2020}

\usepackage[utf8]{inputenc} 
\usepackage[T1]{fontenc}    
\usepackage{hyperref}       
\usepackage{url}            
\usepackage{booktabs}       
\usepackage{amsfonts}       
\usepackage{nicefrac}       
\usepackage{microtype}      
\usepackage{amsmath}
\usepackage{multirow}
\usepackage{adjustbox}

\usepackage{comment}

\usepackage{tikz}
\usetikzlibrary{calc}
\usetikzlibrary{shapes.geometric, arrows, fit}
\usepackage{varwidth}
\usetikzlibrary{positioning}
\usetikzlibrary{backgrounds}
\usetikzlibrary{decorations.pathreplacing}

\definecolor{navy}{RGB}{0,0,137}
\definecolor{tealDeer}{RGB}{148,232,180}
\definecolor{dodgerBlue}{RGB}{18,161,255}
\definecolor{citrine}{RGB}{230, 194, 8}
\definecolor{violet}{RGB}{112,5,164}
\definecolor{navyPurple}{RGB}{172,86,253}
\definecolor{heliotrope}{RGB}{236,93,253}
\definecolor{pink}{RGB}{250,38,160}

\DeclareMathOperator{\vect}{vec}
\newcommand{\R}{\mathbb{R}}

\title{Convolutional Neural Networks from Image Markers}

\author{%
  B\'arbara C. Benato\thanks{BEYOND BACKPROPAGATION: Novel Ideas for Training Neural Architectures - Workshop at NeurIPS, 12th of December 2020},\hspace{1em} Italos S. Estilon, \hspace{1em}Felipe L. Galv\~ao, \hspace{1em} Alexandre X. Falc\~ao \\

   Laboratory of Image Data Science\\
   Institute of Computing \\
  University of Campinas\\
  Campinas, Brazil \\
  \texttt{\{barbara.benato, italos.souza, felipe.galvao, afalcao\}@ic.unicamp.br} \\
}

\begin{document}

\maketitle

\begin{abstract}
 A technique named \textit{Feature Learning from Image Markers} (FLIM) was recently proposed to estimate convolutional filters, with no backpropagation, from strokes drawn by a user on very few images (e.g., 1-3) per class, and demonstrated for coconut-tree image classification. This paper extends FLIM for fully connected layers and demonstrates it on different image classification problems. The work evaluates marker selection from multiple users and the impact of adding a fully connected layer. The results show that FLIM-based convolutional neural networks can outperform the same architecture trained from scratch by backpropagation. 
\end{abstract}

\section{Introduction}
\begin{figure}[!ht]
    \begin{center}
    \begin{tabular}{ccc}
    \includegraphics[width=0.25\hsize]{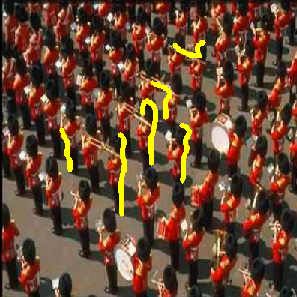}&
    \includegraphics[width=0.25\hsize]{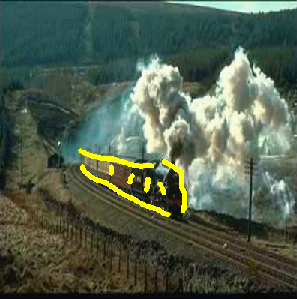}&
    \includegraphics[width=0.25\hsize]{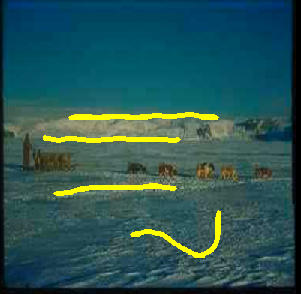}\\
    (a) & (b) & (c)
    \end{tabular}
    \end{center}
    \caption{\footnotesize Marker selection (yellow) on images of three out of six classes, as defined for this work from a Corel Stock CD with JPEG images: (a) royal guard, (b) train, and (c) snow.}  
    \label{fig:markerselection}
\end{figure}

Convolutional neural networks (CNNs) have shown remarkable performance in image classification problems~\cite{wang2019development, zhang2019survey}, mostly due to their capability of extracting relevant features by convolutional layers. On the other hand, CNNs may present complex and deep architectures, challenging their training from scratch, requiring considerable human effort in data annotation, and resulting in non-explainable models. A better understanding of CNNs towards the construction of explainable models has been investigated in several works~\cite{arrieta2020explainable,fukui2019attention,montavon2018methods}. In the same context, user involvement seems crucial to discover more efficient and effective ways to transfer human knowledge to machines during the deep learning process. One example is a recent technique, named \textit{Feature Learning from Image Markers} (FLIM), which can estimate relevant filters to compose a given number of convolutional layers from strokes drawn by a user on very few images (e.g., 1-3) per class~\cite{estilon2020learning}. The strokes are drawn on image regions that best represent the classes, which can be visually identified in many applications. Figure~\ref{fig:markerselection} shows examples with edge and texture regions that characterize distinct image classes. 

This paper extends FLIM for fully connected layers and evaluates marker selection from multiple users in different image classification problems, achieving very interesting results. First, we propose flattening, z-score normalization, and image vector clustering per class to identify a few representative images (cluster centers) for marker selection given an image training set. From each convolutional layer's input, FLIM applies marker-based normalization, extracts patches from the image markers, and performs patch clustering to estimate the convolutional layer's filters as cluster centers. Figure~\ref{fig:flim-conv-layer} illustrates this training process in which a convolutional layer may consist of \textit{marker-based normalization}, convolution, ReLU activation, and max pooling. By flattening the last convolutional layer's output and using the entire training set, we also propose z-score normalization and image vector clustering per class at the input of each fully connected layer to estimate its neurons as cluster centers. The details are explained next.

\section{Feature Learning from Image Markers}
\label{s.FLIM}

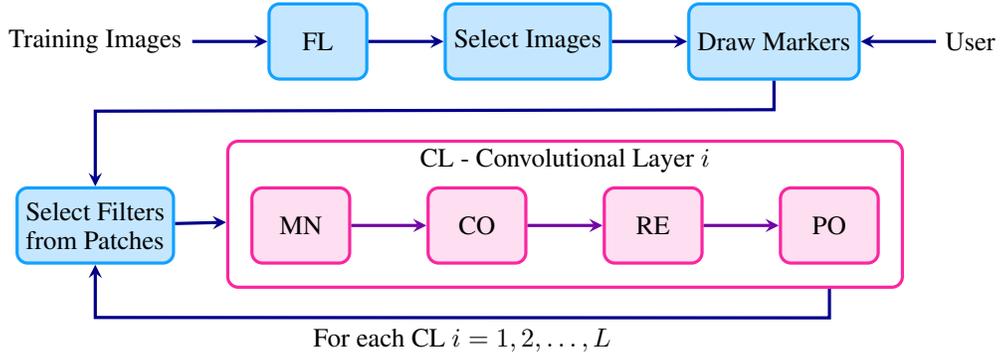
\begin{figure}
  \begin{center}
  
    \tikzstyle{steps} = [rectangle, rounded corners, minimum width=1.3cm, minimum height=1cm,text centered, draw=dodgerBlue, very thick, fill=dodgerBlue!25, execute at begin node={\begin{varwidth}{2.5cm}},
      execute at end node={\end{varwidth}}]
  
    \tikzstyle{myarrow} = [navy, very thick]
    \tikzstyle{convarrow} = [violet, very thick]
  
    \tikzstyle{arrow} = [thick,->,>=stealth]

    \begin{tikzpicture}
      \node (images) [steps,draw=none, fill=white] {Training Images};

      \node (fl) [steps, right= of images] {FL};
      \node (select) [steps, right= of fl] {Select Images};
      \node (draw) [steps, right= of select] {Draw Markers};

      \node (user) [draw=none, fill=white, right= of draw] {User};

      \node (patches) [steps, below=4em of images] {Select Filters \\ from Patches};

      \node (mn) [steps, draw=pink, fill=pink!15, right= of patches] {MN};
      \node (co) [steps, draw=pink, fill=pink!15, right= of mn] {CO};
      \node (re) [steps, draw=pink, fill=pink!15, right= of co] {RE};
      \node (po) [steps, draw=pink, fill=pink!15, right= of re] {PO};

      \node (cl) [fit={($(mn.north west)+(-0.3,0.6)$) ($(po.south east)+(0.3,-0.3)$)}, inner sep=0pt, draw=pink, rounded corners, very thick, label = {[label distance = -1.6em]above:{CL - Convolutional Layer $i$}}] {};
    
      \draw [arrow, myarrow] (images) -- (fl);
      \draw [arrow, myarrow] (fl) -- (select);
      \draw [arrow, myarrow] (select) -- (draw);

      \draw [arrow, myarrow] (user) -- (draw);

      \draw [arrow, myarrow] (draw.south) --++ (0.0em, -1.1em)  -- ($(patches.north)+(0.0,1)$) -| (patches.north);

      \draw [arrow, myarrow] (patches) -- (cl);
      \draw [arrow, convarrow] (mn) -- (co);
      \draw [arrow, convarrow] (co) -- (re);
      \draw [arrow, convarrow] (re) -- (po);

      \draw [arrow, myarrow] ([xshift=10em]cl.south) --++ (0.0em, -1.1em)  -- node [black, below] {For each CL $i = 1, 2, \ldots, L$} ($(patches.south)+(0.0,-2em)$) -| (patches.south);  
  \end{tikzpicture}
  \end{center}
  \caption{\footnotesize Building one convolutional layer after another with FLIM. FL stands for flattening, MN for \textit{marker-based normalization}, CO for convolution, RE for ReLU activation, and PO for max pooling.}
  \label{fig:flim-conv-layer}
\end{figure}

In FLIM, each convolutional layer is trained at once from strokes (Figure~\ref{fig:markerselection}) drawn by a user on very few input images (e.g., 1-3) per class. Each convolutional layer $i=1,2,\ldots,L$ may consist of \textit{marker-based normalization}, convolution with a set of filters $\mathcal{F}_i$, ReLU activation, and max pooling, for instance (Figure~\ref{fig:flim-conv-layer}). The first layer has its filters learned from the selected images' features. From the second layer on, FLIM uses the output from the previous layer. One can maintain convolution and pooling with strides one to preserve markers at the same resolution given the small number of selected images. Then, the desired stride per convolutional layer is only adopted when extracting features from the entire set $\mathcal{D}$ of images $I$, with $m$ channels each, for the purpose of training fully connected and decision layers (i.e., the image classifier).

For a filter $F$ with shape $k \times k \times m$, the convolution between an image $I$ and $F$ at a pixel $p$ can be seen as the inner product between $\vect(F)$ and $\vect(P_I(p))$, where $P_I(p)$ is a patch with shape $k \times k \times m$ around pixel $p$, and $\vect$ is the vectorization operation. Thus $\vect(F)$ is normal to a hyperplane at the origin of~$\R^{k\times k \times m}$, such that local patterns given by $\vect(P_I(p))$ lie on the hyperplane's positive side when they match the pattern filtered by $F$, or lie on the negative side otherwise. A filter $F$ suitable for a given class should encode its local patterns differently from other classes, making $F$ a representative patch for that class. A class may then be represented by several patches identified as cluster centers from image markers. As explained next, \textit{marker-based normalization} is crucial to ensure suitable filtering without bias estimation. It uses mean $\mu\in \R^m$ and standard deviation $\sigma\in \R^m$ extracted from pixel values at drawn markers, subtracting $\mu$ from each image $I \in \mathcal{D}$ and dividing by $\sigma$.

Given a classification problem with $c$ classes, let~$\mathcal{M}_I$ be a set of markers drawn on an image~$I \in \mathcal{D}$ from a class~$l \in \{1,2, \ldots, c\}$. Let $P_l$ be the set of all patches around pixels of $\mathcal{M}_I$ and~${\mathcal{P} = \bigcup_{l \in 1,2, \ldots c}{\mathcal{P}_l}}$ be the set of all patches from markers drawn on images from all classes. Due to marker-based normalization, $\mathcal{P}$ is centralized at the origin of $\R^{k\times k\times m}$. By finding a given number of clusters in each set $\mathcal{P}_l$, one can use their centers to compose the subset of filters in $\mathcal{F}_i$ suitable to extract local patterns of class $l$. We also enforce unit norm to those filters, avoiding preferences among them.

After training the convolutional layers, they are used to extract features from all images in ${\cal D}$. By flattening the output of the last convolutional layer, one can add a fully connected layer by applying z-score normalization and clustering per class to estimate the weight vectors of its neurons as cluster centers. Similarly, z-score normalization eliminates bias in fully connected layers. Additional fully connected layers should follow the same procedure, which has also been applied to the vectorization of input images for the purpose of selecting images to draw markers. 

Next, we use support vector machines~\cite{SVM} as decision layer and compare FLIM-based CNNs with a same network architecture trained by backpropagation.  

\section{Experiments and results}

We used $k$-means for clustering in all situations and selected three datasets to evaluate FLIM using markers drawn by two users: Citrus Leaves~\cite{Data:Citrus:2019} ($604$ images); a subset with $355$ images from a CorelStock CD; and Rock, Paper, and Scissors (RPS)~\cite{Data:RockPaperScissors:2019} with $2892$ images. All images have been rescaled to $400\times 400$ pixels. Each dataset $Z$ was randomly partitioned by stratified sampling into sets $Z_1$ and $Z_2$ with $30\%$ of the samples used for training and $70\%$ for testing, respectively. From $Z_1$, we obtain a very small subset $Z_s \subset Z_1$ of selected images to draw markers. Table~\ref{t.splits} describes the main characteristics of these sets. Note that, simpler is the dataset less images are required for marker selection. 

A drawing tool -- such as a free-hand brush -- was used by each user, $A$ and $B$, to draw strokes on images of $Z_s$. While user $A$ drew strokes on edge and texture regions that characterize each class (Figure~\ref{fig:markerselection}), user $B$ also selected markers on other regions, such as the train smoke in Figure~\ref{fig:markerselection}b, which has similar texture to clouds that might appear in other classes (e.g, class snow). 

After training convolutional layers from $Z_s$, they extract features from $Z_1$ and $Z_2$, and a support vector machine (SVM) classifier~\cite{SVM} is trained on $Z1$, as a decision layer, and used to classify images from $Z_2$. We used the one \emph{vs} one strategy, with $C=100$, and $\sigma = \frac{1}{n}$, for $n$ features, to train the SVM decision layer. The experiments were repeated three times with different sets $Z_s \subset Z_1$, $Z_1$, and $Z_2$. As baseline, we trained by backpropagation a same CNN architecture of each FLIM-based CNN with fully connected layer (the one with the best result between users), starting from random weights (Xavier initialization) and using learning rate $0.001$ and weight decay $0.001$. The learning rate was updated with multiplication factor $0.1$ at every $15$ epochs. We set early stop when the training accuracy reached~$99\%$ on $Z_1$, which happened in all cases. Note that, while FLIM trains convolutional layers on $Z_s$, the baseline uses the entire $Z_1$ to learn its feature extractor. 

We evaluated FLIM-based CNNs with one and two convolutional layers (except RPS which required a single convolutional layer), and with one additional fully connected layer, followed by the SVM decision layer. Note that, FLIM might find a different number of filters per layer at each split, depending on the set of markers, but the other architecture's parameters were fixed (see Table~\ref{t.param}). The percentage of neurons (clusters) per class used for the fully connected layer was different for each split and dataset. It was chosen by each user to verify if the additional fully connected layer could improve the results already obtained with the output of the second convolutional layer. However, such parameter could have been optimized on $Z_1$.

\begin{table*}[!tb]
    \begin{minipage}{.5\textwidth}
      \setlength{\tabcolsep}{2pt}
      \centering
      \footnotesize 
      \caption{\footnotesize Number of samples in $Z_s$, $Z_1$, $Z_2$, and $Z$ sets and the percentage of $Z_s$ in $Z$ (fourth column).}
      \label{t.splits}
      \begin{tabular}{lcrrrrr}
      dataset                            & classes & \multicolumn{1}{c}{$\mid Z_s \mid$} & \multicolumn{1}{c}{$\%$ of $|Z|$} & \multicolumn{1}{c}{$\mid Z_1 \mid$} & \multicolumn{1}{c}{$\mid Z_2 \mid$} & \multicolumn{1}{c}{$\mid Z \mid$} \\ \hline \hline
      Corel & 6 & 13 & 3.70$\%$ & 104 & 251 & 355 \\
      Citrus & 5 & 9 & 1.50$\%$ & 179 & 425 & 604 \\
      RPS & 3 & 6 & 0.02$\%$ & 867 & 2025 & 2892 \\ \hline
      \end{tabular}
    \end{minipage}%
    \qquad
    \begin{minipage}{.4\textwidth}
      \setlength{\tabcolsep}{2pt}
      \footnotesize 
      \centering
      \caption{\footnotesize Fixed parameters in the FLIM-based CNNs: the size $k\times k$ of the filters in all convolutional layers, a given number $f_m$ of filters (clusters) per marker, a pooling size $poolsize$ per layer, and the strides for the first convolutional layer (\emph{st}-$i_1$) and for the second one (\emph{st}-$i_2$).}
      \label{t.param}
      \begin{tabular}{lcrrrr}
      dataset                            & $k\times k$ & $f_m$ & \multicolumn{1}{c}{\emph{poolsize}} & \multicolumn{1}{c}{\emph{st}-$i_1$} & \multicolumn{1}{c}{\emph{st}-$i_2$} \\ \hline \hline
      Corel & 5$\times$5 & 8 & 7$\times$7 & 4 & 2 \\
      Citrus & 3$\times$3 & 8 & 7$\times$7 & 4 & 2 \\
      RPS & 3$\times$3 & 8 & 7$\times$7 & 4 & - \\ \hline
      \end{tabular}
    \end{minipage} 
\end{table*}

Table~\ref{t.results1} shows the classification results (accuracy) per split, dataset, and user ($A$ and $B$). The users were experts in image processing and machine learning.  The results of the baseline trained by backpropagation are indicated by $bp$.

\begin{table*}[tb]
  \centering
  \footnotesize
  \caption{\footnotesize Accuracy results for each split, dataset, and user, $A$ and $B$, using FLIM-based CNNs with one ($CL1$) and two ($CL2$) convolutional layers, and one additional fully connected ($FC$) layer. Similarly, for the baseline CNN trained by backpropagation ($bp$). Best results are in bold and the mean results are presented on the right.}
  \setlength{\tabcolsep}{3pt}
  \begin{adjustbox}{width=\textwidth}
  \label{t.results1}
  \begin{tabular}{l|l|rrr|rrr|rrr|rrr}
   &
     &
    \multicolumn{3}{c|}{split1} &
    \multicolumn{3}{c|}{split2} &
    \multicolumn{3}{c|}{split3} &
    \multicolumn{3}{c}{mean} \\
  database &
     &
    \multicolumn{1}{c}{$CL1$} &
    \multicolumn{1}{c}{$CL2$} &
    \multicolumn{1}{c|}{$FC$} &
    \multicolumn{1}{c}{$CL1$} &
    \multicolumn{1}{c}{$CL2$} &
    \multicolumn{1}{c|}{$FC$} &
    \multicolumn{1}{c}{$CL1$} &
    \multicolumn{1}{c}{$CL2$} &
    \multicolumn{1}{c|}{$FC$} &
    \multicolumn{1}{c}{$CL1$} &
    \multicolumn{1}{c}{$CL2$} &
    \multicolumn{1}{c}{$FC$} \\ \hline \hline
  \multirow{3}{*}{Corel} &
    $A$ &
    0.9084 &
    \textbf{0.9203} &
    0.9004 &
    0.9363 &
    \textbf{0.9442} &
    0.9203 &
    0.9004 &
    0.8964 &
    0.8964 &
    0.9150 &
    \textbf{0.9203} &
    0.9057 \\
   &
    $B$ &
    0.8884 &
    0.9043 &
    0.8844 &
    0.9322 &
    0.9362 &
    0.9083 &
    0.9043 &
    0.9163 &
    \textbf{0.9243} &
    0.9083 &
    0.9189 &
    0.9057 \\
   &
    $bp$ &
    \multicolumn{1}{c}{-} &
    \multicolumn{1}{c}{-} &
    0.8690 &
    \multicolumn{1}{c}{-} &
    \multicolumn{1}{c}{-} &
    0.8460 &
    \multicolumn{1}{c}{-} &
    \multicolumn{1}{c}{-} &
    0.8170 &
    \multicolumn{1}{c}{-} &
    \multicolumn{1}{c}{-} &
    0.8438 \\ \hline
  \multirow{3}{*}{Citrus} &
    $A$ &
    0.7859 &
    0.8118 &
    \textbf{0.8400} &
    0.7671 &
    0.7718 &
    \textbf{0.8306} &
    0.7859 &
    0.7788 &
    0.8024 &
    0.7796 &
    0.7875 &
    0.8243 \\
   &
    $B$ &
    0.7317 &
    0.7764 &
    0.8282 &
    0.7858 &
    0.7976 &
    0.8258 &
    0.8000 &
    0.8305 &
    \textbf{0.8352} &
    0.7725 &
    0.8015 &
    \textbf{0.8298} \\
   &
    $bp$ &
    \multicolumn{1}{c}{-} &
    \multicolumn{1}{c}{-} &
    0.7110 &
    \multicolumn{1}{c}{-} &
    \multicolumn{1}{c}{-} &
    0.6870 &
    \multicolumn{1}{c}{-} &
    \multicolumn{1}{c}{-} &
    0.7090 &
    \multicolumn{1}{c}{-} &
    \multicolumn{1}{c}{-} &
    0.7022 \\ \hline
  \multirow{3}{*}{\begin{tabular}[c]{@{}l@{}}RPS\end{tabular}} &
    $A$ &
    \textbf{0.9891} &
    \multicolumn{1}{c}{-} &
    0.9877 &
    0.9852 &
    \multicolumn{1}{c}{-} &
    \textbf{0.9896} &
    0.9921 &
    \multicolumn{1}{c}{-} &
    \textbf{0.9965} &
    0.9888 &
    \multicolumn{1}{c}{-} &
    \textbf{0.9913} \\
   &
    $B$ &
    0.9827 &
    \multicolumn{1}{c}{-} &
    0.9866 &
    0.9767 &
    \multicolumn{1}{c}{-} &
    \textbf{0.9896} &
    0.9906 &
    \multicolumn{1}{c}{-} &
    0.9911 &
    0.9833 &
    \multicolumn{1}{c}{-} &
    0.9891 \\
   &
    $bp$ &
    \multicolumn{1}{c}{-} &
    \multicolumn{1}{c}{-} &
    0.9750 &
    \multicolumn{1}{c}{-} &
    \multicolumn{1}{c}{-} &
    0.9790 &
    \multicolumn{1}{c}{-} &
    \multicolumn{1}{c}{-} &
    0.9770 &
    \multicolumn{1}{c}{-} &
    \multicolumn{1}{c}{-} &
    0.9771 \\ \hline
  \end{tabular}
  \end{adjustbox}
 \end{table*}
 
One can observe that the FLIM-based CNNs (with and without fully connected layer) outperformed the baseline CNN trained by backpropagation ($bp$) in all datasets, splits, and on average. This demonstrates the effectiveness of learning features from image markers, and with no backpropagation, which is a very interesting result. The addition of a fully connected layer was not an advantage in Corel, but in most cases (datasets, splits, and on average) the fully connected layer could improve the best result of the FLIM-based network with convolutional layers only. This indicates that the proposed approach, that computes z-score normalization and clustering to identify selected images and neurons for fully connected layers, is indeed effective. When comparing the performance of marker selection from users $A$ and $B$, the strokes on representative parts of the classes seem to be the best strategy, as followed by user $A$, but the additional selection of markers on parts that do not represent classes, as followed by user $B$, was better in some cases (e.g., split 3 in Citrus and Corel). Although this might happen, we believe that markers should always indicate regions that represent classes in order to avoid CNNs that learn to solve a dataset but not the image classification problem.  

\section{Conclusions}

This paper investigates a recently proposed approach, FLIM, to learn features from strokes drawn by users on very few images per class. We proposed a method to select images from the input training set to draw markers and to select neurons (their weight vectors) for fully connected layers. The experiments with two users and using three datasets with distinct image properties have demonstrated that: (i) it is possible to effectively train CNNs from image markers on representative regions of the classes with no backpropagation, (ii) FLIM-based CNNs can outperform a same CNN architecture trained by backpropagation, and (iii) a fully connected layer may improve the results of the FLIM-based CNN with convolutional layers only.

We must say that not only the number of selected images was small but also the number of markers per image was small (e.g., less than 10 markers). As pros, FLIM allows an intuitive, straightforward, and explainable mechanism to train CNNs with reduced user effort. As cons, for problems with many classes, marker selection in all classes requires more user effort. It is possible that markers selected in some classes result filters that can extract useful features from other classes, but we do not know that yet. We intend to investigate other alternatives for clustering since the method is responsible to select images, filters, and neurons. We also intend to verify the impact of increasing the number of selected images, markers, datasets, and their sizes.

\bibliographystyle{abbrv}
\bibliography{refs}

\end{document}